\documentclass{article}

\usepackage[final]{neurips_2024}

\usepackage[utf8]{inputenc} 
\usepackage[T1]{fontenc}    
\usepackage{hyperref}       
\usepackage{url}            
\usepackage{booktabs}       
\usepackage{amsfonts}       
\usepackage{nicefrac}       
\usepackage{microtype}      
\usepackage{xcolor}         
\usepackage{graphicx}
\usepackage{subfigure}
\usepackage{amsmath} 
\usepackage{float}    
\usepackage{array}

\title{
  From Imitation to Optimization: A Comparative Study of Offline Learning for Autonomous Driving
  }

\author{%
  Antonio Guillen-Perez  \\
  Independent Researcher \\
  \texttt{antonio\_algaida@hotmail.com} \\
}

\begin{document}

\maketitle

\begin{abstract}
  Learning robust driving policies from large-scale, real-world datasets is a central challenge in autonomous driving, as online data collection is often unsafe and impractical. While Behavioral Cloning (BC) offers a straightforward approach to imitation learning, policies trained with BC are notoriously brittle and suffer from compounding errors in closed-loop execution. This work presents a comprehensive pipeline and a comparative study to address this limitation. We first develop a series of increasingly sophisticated BC baselines, culminating in a Transformer-based model that operates on a structured, entity-centric state representation. While this model achieves low imitation loss, we show that it still fails in long-horizon simulations. We then demonstrate that by applying a state-of-the-art Offline Reinforcement Learning algorithm, Conservative Q-Learning (CQL), to the same data and architecture, we can learn a significantly more robust policy. Using a carefully engineered reward function, the CQL agent learns a conservative value function that enables it to recover from minor errors and avoid out-of-distribution states. In a large-scale evaluation on 1,000 unseen scenarios from the Waymo Open Motion Dataset, our final CQL agent achieves a \textbf{3.2x higher success rate} and a \textbf{7.4x lower collision rate} than the strongest BC baseline, proving that an offline RL approach is critical for learning robust, long-horizon driving policies from static expert data.
\end{abstract}


\section{Introduction}
The development of safe and reliable autonomous vehicles relies heavily on the ability to learn complex decision-making policies from real-world driving data. Large-scale datasets, such as the Waymo Open Motion Dataset (WOMD) \cite{Ettinger2021}, provide an invaluable resource, offering millions of examples of expert human driving in diverse and challenging scenarios. However, learning directly from this static, "offline" data presents a significant challenge.

A common approach is Behavioral Cloning (BC), where a policy is trained via supervised learning to mimic the expert's actions. While simple and often effective for one-step prediction, BC policies are known to be brittle in real-world, closed-loop deployment. Minor prediction errors can accumulate over time, a phenomenon known as \textbf{covariate shift} \cite{Ross2011}, pushing the agent into unfamiliar states where its learned policy is unreliable, often leading to catastrophic failure.

This paper investigates a more robust paradigm: Offline Reinforcement Learning (Offline RL). Unlike BC, Offline RL aims to learn a value function that estimates the long-term, goal-oriented outcome of actions, allowing the agent to make intelligent decisions even when it deviates from the expert's exact trajectory. We present an end-to-end pipeline for applying and rigorously evaluating these two learning paradigms on the WOMD. Our key contributions are:
\begin{itemize}
    \item A complete, open-source pipeline for processing raw WOMD data into a structured, ML-ready format, including a parallelized feature engineering process.
    \item A thorough, comparative evaluation of multiple Behavioral Cloning architectures, from simple MLPs to a state-of-the-art Transformer-based model, demonstrating that architectural improvements alone do not solve the fundamental compounding error problem.
    \item A successful application of the Conservative Q-Learning (CQL) algorithm, demonstrating that it learns a significantly more robust and successful driving policy than the best imitation learning baseline on the same data.
\end{itemize}
Through quantitative and qualitative results, we show that the value-based, conservative regularization of Offline RL is a critical component for learning robust, long-horizon policies from large-scale, offline datasets. The complete source code, including all scripts for data processing, training, and evaluation, as well as the final trained model weights, are made publicly available in our GitHub repository: \url{https://github.com/AntonioAlgaida/WaymoOfflineRL}.


\section{Related Work}

This work builds upon three core areas of research: imitation learning for autonomous control, the field of offline reinforcement learning, and the development of large-scale public datasets and simulators.

\paragraph{Imitation Learning and its Limitations.}
Imitation Learning, particularly Behavioral Cloning (BC), represents a foundational and widely-used paradigm for learning control policies from expert demonstrations \cite{Pomerleau1989, Bain1995}. By framing the control problem as a standard supervised learning task, BC aims to learn a direct mapping from states to expert actions. Researchers have applied increasingly powerful function approximators, from early MLPs to modern Transformer architectures, to minimize this one-step prediction error \cite{Codevilla2019}. However, a well-documented and fundamental limitation of BC is the problem of \textbf{covariate shift} \cite{Ross2010}. As first systematically analyzed by Ross et al., small, compounding errors in the learned policy lead the agent into out-of-distribution states that were not present in the expert's dataset. From these unfamiliar states, the policy's predictions become unreliable, often leading to a catastrophic failure spiral. Interactive methods like DAgger \cite{Ross2011} were proposed to mitigate this by querying an expert during training, but such methods are not applicable in a purely offline setting. Our work explicitly demonstrates the covariate shift failure mode, even with a sophisticated Transformer-based policy, motivating the need for a learning paradigm that can reason about the consequences of its actions without an interactive expert.

\paragraph{Offline Reinforcement Learning.}
Offline Reinforcement Learning (Offline RL), also known as Batch RL, is designed to overcome the limitations of BC by learning a value function and policy from a fixed, static dataset without further interaction with the environment \cite{Lange2012, Levine2020}. This makes it particularly well-suited for real-world applications like driving, where online exploration is unsafe and costly. A key challenge in Offline RL is \textbf{extrapolation error}, where standard off-policy value-based methods like Q-learning can become overly optimistic about the value of unseen, out-of-distribution actions, leading to divergent and poor policies \cite{Fujimoto2019}. Modern Offline RL algorithms address this by introducing a notion of conservatism, either through policy constraints or value regularization. Our work leverages \textbf{Conservative Q-Learning (CQL)} \cite{Kumar2020}, a state-of-the-art, value-regularization algorithm. CQL learns a conservative Q-function by adding a penalty term to its objective that explicitly minimizes the Q-values for out-of-distribution actions while pushing up the values for actions found in the dataset. By applying CQL, we aim to learn a policy that is robust to the distributional shift that causes BC to fail, enabling it to recover from minor errors and complete long-horizon tasks.

\paragraph{Datasets and Simulators for Autonomous Driving Research.}
Progress in data-driven methods for autonomous driving is critically dependent on the availability of large-scale, high-fidelity public datasets. The \textbf{Waymo Open Motion Dataset} \cite{Ettinger2021}, which we use in this work, is a massive and influential dataset providing synchronized trajectories and detailed map information for thousands of complex, real-world driving scenarios. For evaluation, traditional high-fidelity simulators like CARLA \cite{Dosovitskiy2017} are invaluable. However, for the large-scale, parallelized evaluation required by modern research, lightweight, JAX-based simulators have emerged. We use the \textbf{Waymax} simulator \cite{Gilles2023} for our closed-loop evaluation due to its high performance and direct compatibility with the Waymo dataset, enabling efficient and reproducible testing of our learned policies.


\section{Methodology}
\label{sec:methodology}

Our methodology is structured as an iterative progression, starting with a robust imitation learning baseline and culminating in a state-of-the-art offline reinforcement learning agent. Each stage is designed to address the limitations of the previous one. All spatial features are represented in an ego-centric reference frame, where the self-driving car (SDC) is at the origin and facing forward.

\subsection{Problem Formulation as a Markov Decision Process}
\label{sec:mdp}

We formulate the autonomous driving task as a discrete-time, continuous state and action space Markov Decision Process (MDP). An MDP is defined by a tuple $(\mathcal{S}, \mathcal{A}, \mathcal{P}, \mathcal{R}, \gamma)$, where:
\begin{itemize}
    \item $\mathcal{S}$ is the set of all possible states. A state $s_t \in \mathcal{S}$ is the structured, ego-centric representation of the driving scene at timestep $t$, as detailed in Section~\ref{sec:state_representation}.
    \item $\mathcal{A}$ is the set of all possible actions. An action $a_t \in \mathcal{A}$ is the 2D kinematically-plausible control vector $(\text{acceleration}, \text{steering angle})$, as detailed in Section~\ref{sec:action_representation}.
    \item $\mathcal{P}(s_{t+1} | s_t, a_t)$ is the state transition probability function, which is unknown to the agent and is implicitly defined by the complex dynamics of the multi-agent traffic environment.
    \item $\mathcal{R}(s_t)$ is the reward function, which provides a scalar feedback signal indicating the desirability of being in a given state. Our design is detailed in Section~\ref{sec:reward}.
    \item $\gamma$ is the discount factor, which balances the importance of immediate versus future rewards.
\end{itemize}
The agent's goal is to learn a policy $\pi(a_t | s_t)$ that maximizes the expected cumulative discounted reward, or return: $\mathbb{E}_{\pi} \left[ \sum_{t=0}^{T} \gamma^t \mathcal{R}(s_t) \right]$. Since we operate in the offline setting, the policy must be learned from a fixed, pre-collected dataset of expert trajectories, $\mathcal{D} = \{ (s_t, a_t, r_{t+1}, s_{t+1}, d_{t+1}) \}$.

\subsection{State and Action Representation}
\label{sec:state_representation}

\begin{figure}[t] 
  \centering
  \includegraphics[width=0.8\textwidth]{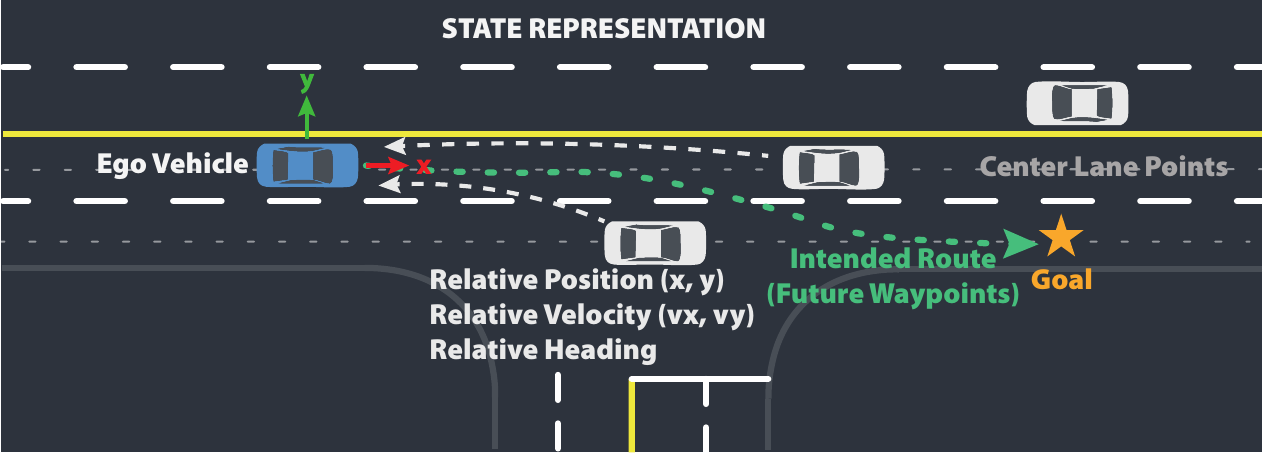} 
  \caption{
    An illustration of the structured, ego-centric state representation. The agent (SDC) perceives the world from its own reference frame (X-axis forward, Y-axis left). It receives information about its own kinematics (\textbf{Ego State}), the relative states of nearby vehicles (\textbf{Agent States}), the geometry of the road (\textbf{Map Features}), and its intended path (\textbf{Route and Goal}). This structured input allows the policy to learn a rich, contextualized understanding of the driving scene.
  }
  \label{fig:state_representation}
\end{figure}

A robust driving policy requires a comprehensive understanding of its environment, its own physical state, and its intended goal. We designed a rich, structured state representation that is provided to our policy network as a dictionary of entity sets. This preserves the inherent structure of the scene and allows our models to reason about different entity types independently. All spatial features are transformed into an ego-centric reference frame, as illustrated in Figure~\ref{fig:state_representation}.

\paragraph{State Representation.}
The state dictionary $s_t$ contains the following key-value pairs, where each value is a tensor:
\begin{itemize}
    \item \textbf{Ego State ($s_t[\text{ego}]$):} A 3-dimensional vector describing the SDC's immediate kinematics. It contains $[\text{speed (m/s)}, \text{acceleration (m/s}^2\text{)}, \text{yaw rate (rad/s)}]$. These values are computed from the SDC's simulated trajectory, providing a true representation of its current motion.

    \item \textbf{Agent States ($s_t[\text{agents}]$):} A tensor of shape $(N, 10)$ representing the $N=15$ closest agents. Each agent is described by a 10-dimensional feature vector: $[x, y, v_x, v_y, \text{heading}, \text{length}, \text{width}, \text{is\_vehicle}, \text{is\_pedestrian}, \text{is\_cyclist}]$. All features are relative to the SDC's frame. This fixed-size tensor, created by sorting all scene agents by distance and padding with zeros if fewer than $N$ are present, provides a consistent input structure for the policy network.

    \item \textbf{Map Geometry ($s_t[\text{lanes}]$, $s_t[\text{crosswalks}]$):} Tensors representing the local road geometry. 
        \begin{itemize}
            \item $s_t[\text{lanes}]$ (shape $(M, 2)$): The ego-centric $(x,y)$ coordinates of the $M=50$ closest lane centerline points.
            \item $s_t[\text{crosswalks}]$ (shape $(K, 2)$): The ego-centric $(x,y)$ coordinates of the $K=10$ closest crosswalk polygon points.
        \end{itemize}
    This representation provides the model with a point-cloud-like understanding of the immediate road structure.

    \item \textbf{Route and Rules ($s_t[\text{route}]$, $s_t[\text{rules}]$):} Tensors that provide critical goal-directed and rule-based context.
        \begin{itemize}
            \item $s_t[\text{route}]$ (shape $(10, 2)$): The ego-centric $(x,y)$ coordinates of 10 future waypoints sampled from the expert's intended trajectory. This gives the agent explicit, long-horizon information about the desired path.
            \item $s_t[\text{rules}]$ (shape $(8,)$): A vector containing task-critical information: [\text{dist\_to\_goal}, \text{goal\_dir\_x}, \text{goal\_dir\_y}, \text{dist\_to\_stop\_sign}, \text{is\_stop\_controlled}, \text{traffic\_light\_state (one-hot, 3 dims)}]. The traffic light state indicates if the SDC's current lane is controlled by a green, yellow, or red signal.
        \end{itemize}
\end{itemize}

A summary of the states can be seen in Table \ref{tab:state_representation}.

\begin{table}[h]
  \caption{Structure of the State Representation Dictionary ($s_t$). All spatial features are in the SDC's ego-centric frame.}
  \label{tab:state_representation}
  \centering
  \resizebox{\textwidth}{!}{%
  \begin{tabular}{l c >{\raggedright\arraybackslash}p{5.5cm} l}
    \toprule
    \textbf{Key} & \textbf{Tensor Shape} & \textbf{Description} & \textbf{Feature Breakdown (per entity)} \\
    \midrule
    \texttt{ego} & $(3,)$ & SDC's immediate kinematic state. & $[\text{speed}, \text{accel}, \text{yaw\_rate}]$ \\
    \texttt{agents} & $(15, 10)$ & The 15 closest agents, sorted by distance. Padded with zeros. & $[\text{rel}_{x}, \text{rel}_{y}, \text{rel}_{v_x}, \text{rel}_{v_y}, \text{rel}_{h}, \text{len}, \text{wid}, \text{is\_veh}, \text{is\_ped}, \text{is\_cyc}]$ \\
    \texttt{lanes} & $(50, 2)$ & The 50 closest lane centerline points. & $[\text{rel}_{x}, \text{rel}_{y}]$ \\
    \texttt{crosswalks} & $(10, 2)$ & The 10 closest crosswalk polygon points. & $[\text{rel}_{x}, \text{rel}_{y}]$ \\
    \texttt{route} & $(10, 2)$ & 10 future waypoints on the expert's path. & $[\text{rel}_{x}, \text{rel}_{y}]$ \\
    \texttt{rules} & $(8,)$ & Goal-directed and rule-based info. & $[d_{\text{goal}}, \text{dir}_{gx}, \text{dir}_{gy}, d_{\text{stop}}, \text{is\_stop}, \text{tl}_{g}, \text{tl}_{y}, \text{tl}_{r}]$ \\
    \bottomrule
  \end{tabular}
  }
\end{table}

\paragraph{Action Representation.}
\label{sec:action_representation}

The agent's control policy, or \textbf{action} $a_t$, is defined in a physically-plausible kinematic space as a 2-dimensional vector. The components and their physical ranges are detailed in Table~\ref{tab:action_representation}.

\begin{table}[H]
  \caption{Structure of the Kinematic Action Vector ($a_t$).}
  \label{tab:action_representation}
  \centering
  \begin{tabular}{lccl}
    \toprule
    \textbf{Component} & \textbf{Index} & \textbf{Physical Range} & \textbf{Description} \\
    \midrule
    Acceleration & 0 & $[-10.0, 8.0]$ & The target longitudinal acceleration in m/s$^2$. \\
    Steering Angle & 1 & $[-0.8, 0.8]$ & The target front wheel steering angle in radians. \\
    \bottomrule
  \end{tabular}
\end{table}

To obtain the ground-truth expert actions ($a_t^*$) for imitation learning, we leverage the inverse dynamics of the simulation model. Specifically, we use the `compute\_inverse` function from the `InvertibleBicycleModel` provided by the Waymax simulator \cite{Gilles2023}. This function takes the expert's state transition from $s_t$ to $s_{t+1}$ and calculates the precise kinematic action $a_t^*$ that would produce this motion. This ensures perfect consistency between the actions our agent learns to imitate and the physics of the environment in which it will be evaluated. To handle noise and artifacts in the source data, the calculated expert actions are clipped to the physical ranges defined in Table~\ref{tab:action_representation} during our data preprocessing stage. The policy network's output is squashed by a `tanh` activation and then rescaled to ensure its predictions always fall within these valid bounds.

\subsection{Behavioral Cloning Baselines}
\label{sec:bc}

To contextualize the performance of our final RL agent, we first establish a series of increasingly sophisticated baselines using Behavioral Cloning (BC). BC treats policy learning as a supervised regression problem, training a network $\pi(s_t)$ to minimize the Mean Squared Error (MSE) against the expert's ground-truth action $a_t^*$. We implemented three distinct architectures to explore the impact of state representation and model complexity.

\paragraph{BC-K: Kinematic MLP with a Flat State Vector.}
Our simplest baseline, BC-K (Kinematic), uses a standard Multi-Layer Perceptron (MLP) architecture. The input to this model is a single, \textbf{flattened feature vector} of 279 dimensions. This vector is created by concatenating all the engineered features (ego, agents, map, and rules) into one long array. This approach is computationally efficient but destroys the inherent structure of the scene, forcing the model to learn relationships between disparate parts of the input vector from scratch.

\paragraph{BC-S: Structured MLP with Entity Encoders.}
Our second baseline, BC-S (Structured), addresses the limitations of the flat vector by processing the structured state dictionary directly. It uses an \textbf{entity-centric architecture} where separate, smaller MLP encoders first process the features for each entity set (ego, agents, lanes, etc.) to produce fixed-dimensional embeddings. These individual entity embeddings are then aggregated via a permutation-invariant max-pooling operation. The aggregated vectors are concatenated and passed to a final MLP head that outputs the action. This structure allows the model to learn specialized representations for each entity type but aggregates their information in a simple, non-relational way.

\paragraph{BC-T: Transformer-based Policy.}
Our strongest baseline, BC-T (Transformer), replaces the simple max-pooling aggregation with a more powerful \textbf{Transformer Encoder} architecture \cite{Vaswani2017}. The embeddings from the entity encoders are concatenated into a single sequence and processed by a multi-layer Transformer. The self-attention mechanism allows this model to explicitly reason about the relationships and interactions between all entities in the scene. The output token corresponding to the ego-vehicle, which now contains a rich, contextualized summary of the entire scene, is passed to the final MLP head to predict the action. The detailed architecture is identical to that of our final CQL agent and is described in full in Section~\ref{sec:setup_architecture}.

\subsection{Offline Reinforcement Learning with CQL}
\label{sec:cql}

While the BC baselines are effective at one-step prediction, they inevitably fail in closed-loop simulation due to compounding errors. To address this, we train our final agent using Conservative Q-Learning (CQL) \cite{Kumar2020}, a state-of-the-art offline RL algorithm.

\subsection{Reward Function Engineering}
\label{sec:reward}

To enable reinforcement learning, we designed a dense, multi-objective reward function $R(s_t)$ to provide a learning signal at every timestep. The function is a weighted combination of components that define our desired driving behavior. The final score is normalized to a consistent range using a hyperbolic tangent function to ensure stable training. The total reward is calculated as:
\begin{equation}
    R(s_t) = \tanh \left( \frac{1}{C} \left( w_r r_{\text{route}} + w_s p_{\text{safety}} + w_c p_{\text{comfort}} \right) \right)
\end{equation}
where $C$ is a scaling factor and $w$ are the component weights. The individual reward and penalty terms are summarized in Table~\ref{tab:reward_components}. This squashing normalization is crucial for bounding the final reward signal to the range $[-1, 1]$, preventing un-normalized penalty values from creating extreme gradients and destabilizing the critic's learning process.

\begin{table}[H]
  \caption{Components of the Multi-Objective Reward Function.}
  \label{tab:reward_components}
  \centering
  \begin{tabular}{p{0.2\textwidth} p{0.55\textwidth} c}
    \toprule
    \textbf{Component} & \textbf{Description} & \textbf{Signal} \\
    \midrule
    Route Following & A score combining forward speed projected onto the intended route with a penalty for cross-track error. & Reward \\
    \midrule
    Safety & A penalty that increases quadratically as the Time-to-Collision (TTC) with a forward vehicle drops below a safety threshold. & Penalty \\
    \midrule
    Comfort & A penalty for physically uncomfortable maneuvers, based on the sum of squared longitudinal acceleration and squared yaw rate. & Penalty \\
    \bottomrule
  \end{tabular}
\end{table}

\paragraph{CQL Algorithm.}
CQL learns an Actor (policy) and a Critic (Q-function) from our pre-collected, static dataset. The Critic is trained to minimize a modified Bellman error that includes a conservative regularization term. For a batch of transitions $\mathcal{B}$ from the dataset $D$, the simplified objective for the Q-function $\mathcal{Q}$ is:
\begin{equation}
    \min_{\mathcal{Q}} \quad \alpha \left( \mathbb{E}_{s \sim \mathcal{D}} \left[ \log\sum_{a}e^{\mathcal{Q}(s,a)} \right] - \mathbb{E}_{s,a \sim \mathcal{D}} \left[ \mathcal{Q}(s,a) \right] \right) + \mathcal{L}_{\text{Bellman}}(\mathcal{Q})
\end{equation}
The first term is the core of CQL: it pushes down the Q-values for arbitrary actions (`a`) while pushing up the Q-values for actions observed in the dataset (`s,a ~ D`). This makes the learned Q-function "conservative," or pessimistic, about the value of actions outside the expert data distribution. The Actor is then trained to produce actions that maximize this conservative Q-function. This process yields a policy that is robust to distributional shift and can recover from small errors by actively avoiding unfamiliar, low-value actions. Our implementation uses the Transformer-based architecture from BC-T for both the Actor and the Critic.

\section{Experiments}
\label{sec:experiments}

We conduct a series of experiments to evaluate the effectiveness of our learning pipeline. We first train and evaluate several Behavioral Cloning baselines with increasing architectural complexity, and then train our final Offline RL agent and compare its performance.

\subsection{Dataset and Preprocessing Pipeline}
\label{sec:dataset}

All experiments are conducted using the large-scale \textbf{Waymo Open Motion Dataset (WOMD)} \cite{Ettinger2021}. We utilize the v1.1 `scenario` format, which provides rich, vectorized map data and full 9-second (10Hz) agent trajectories. Our dataset consists of 200 training and 50 validation shards, comprising approximately 200,000 unique driving scenarios. To manage this scale, we developed a high-performance, multi-stage data processing pipeline.

\paragraph{1. Parallelized Parsing.}
The raw dataset is provided as sharded `.tfrecord` files containing serialized `Scenario` protocol buffers. We first perform a parallelized parsing stage where multiple worker processes, managed by Python's `multiprocessing` library, read the `.tfrecord` files concurrently. Each worker parses the `Scenario` protos to extract all relevant information---including agent trajectories, object metadata, dynamic traffic light states, and detailed static map features---and saves them into a set of intermediate `.npz` files, with one file per scenario. This initial stage converts the proprietary data format into a universally accessible NumPy-based structure.

\paragraph{2. Feature Engineering and Action Derivation.}
In the second stage, we process the `.npz` files to generate the final training samples. This stage is also heavily parallelized, with each worker responsible for a subset of the scenarios. For each of the 90 possible timesteps within a scenario, we construct the structured state dictionary detailed in Section~\ref{sec:state_representation}. This involves several computationally intensive feature engineering steps:
\begin{itemize}
    \item \textbf{Vectorized Nearest-Neighbor Search:} To efficiently find the closest map and agent features for all 91 timesteps of a scenario at once, we employ a vectorized approach. We construct large distance matrices using NumPy's broadcasting capabilities (e.g., a `(91, num\_agents)` matrix for agent distances) and perform a single `np.argsort` operation to find the closest entities for the entire scenario, avoiding slow, per-timestep loops.
    \item \textbf{Ego-centric Transformation:} All spatial features (agent positions, map points, route waypoints) are transformed into the SDC's local coordinate frame for each timestep.
    \item \textbf{Kinematic Feature Calculation:} The SDC's current state features, such as `acceleration` and `yaw\_rate`, are calculated from its trajectory using finite differences.
    \item \textbf{Route and Rule Inference:} The agent's intended route is extracted from the expert's future trajectory. The state of relevant traffic lights and the applicability of stop signs are inferred by cross-referencing the SDC's determined lane position with the dynamic and static map data.
\end{itemize}
The ground-truth expert action $(\text{acceleration}, \text{steering angle})$ for each timestep is derived using the `compute\_inverse` function of the simulator's `InvertibleBicycleModel` \cite{Gilles2023}. This ensures perfect physical consistency between the training data and the evaluation environment.

\paragraph{3. Data Cleaning and Normalization.}
To create a clean and stable learning problem, we apply two final processing steps. First, to handle noise and artifacts from the inverse dynamics calculation, we \textbf{clip} the expert actions to a pre-defined, physically plausible range (e.g., accelerations in $[-10, 8]$ m/s$^2$). This is critically different from filtering; clipping preserves the full, unbroken sequence of 90 timesteps per scenario, which is essential for our sequential RL dataset. Second, we perform a final parallelized pass over the entire cleaned training dataset to compute the per-feature \textbf{mean and standard deviation}. These statistics are saved and used to normalize all state inputs to have zero mean and unit variance, a crucial step for stable neural network training.

\paragraph{4. Final Data Format.}
The final preprocessed dataset is saved in a \textbf{"scenario-per-file"} format. Each `.pt` file contains a list of all 90 clipped and structured `(state\_dict, action\_tensor, timestep)` samples for a single scenario, in chronological order. This format is architecturally clean and is used as the high-performance data source for both our Behavioral Cloning and Offline RL `Dataset` objects.

\subsection{Experimental Setup}
\label{sec:setup}

We evaluate three primary agent architectures: a simple MLP with entity encoders (BC-S), a Transformer-based policy (BC-T), and our final Offline RL agent (CQL), which also uses the Transformer architecture. All models are trained and evaluated using PyTorch.

\begin{figure}[t]
  \centering
  \includegraphics[width=\textwidth]{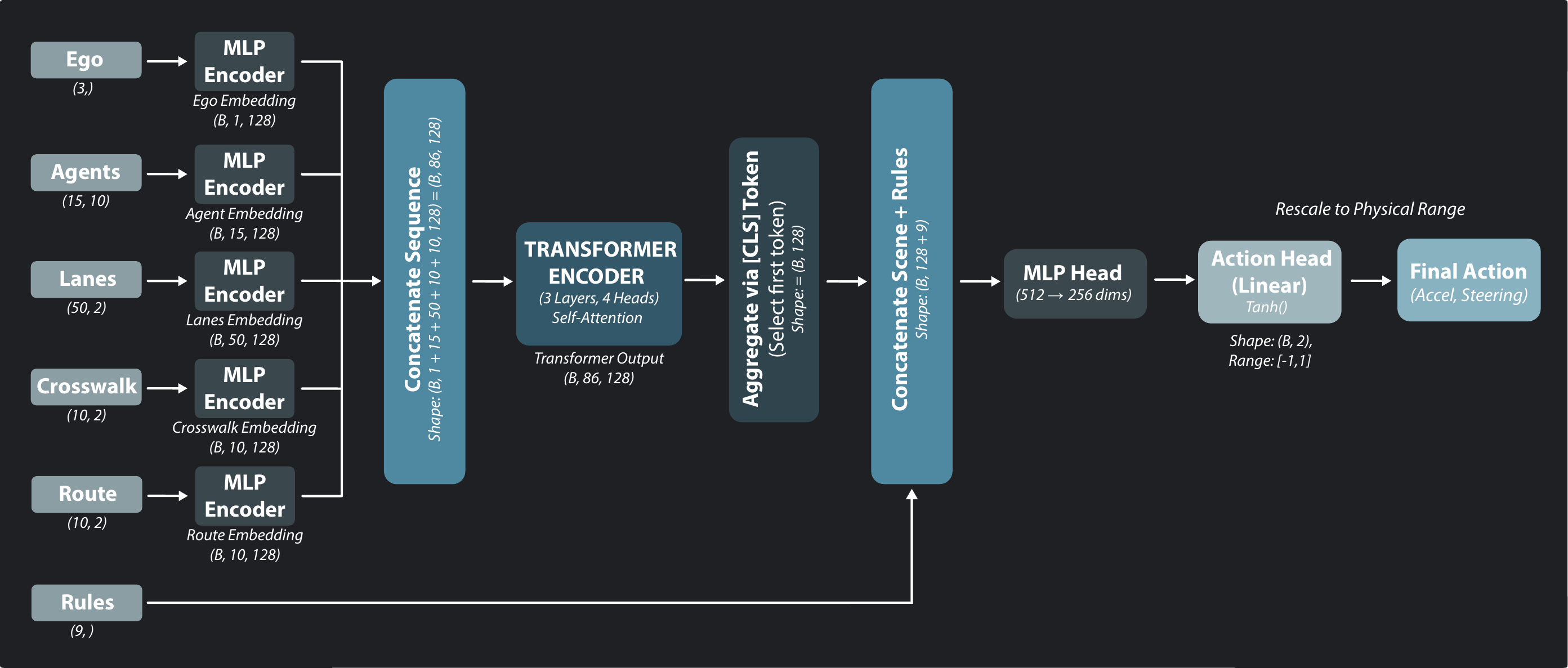} 
  \caption{
    The Transformer-based policy architecture. Raw, ego-centric features for each entity set are first projected into a common embedding space. A Transformer Encoder uses self-attention to reason about the relationships between all entities. The aggregated scene representation is then combined with non-spatial rule features and passed to a final MLP head to produce the kinematic action.
  }
  \label{fig:architecture}
\end{figure}

\paragraph{Network Architecture.}
\label{sec:setup_architecture}
Our primary model architecture for both the final Behavioral Cloning agent (BC-T) and the CQL agent is a Transformer-based network designed to process our structured, entity-centric state representation. The complete data flow of the policy network is illustrated in Figure~\ref{fig:architecture}.

The model consists of five key stages, as shown in the diagram:
\begin{itemize}
    \item \textbf{1. Entity Encoders:} We first use separate, small MLPs to project the raw features for each spatial entity type (ego, agents, lanes, crosswalks, and route) into a shared $D=128$ dimensional embedding space.
    \item \textbf{2. Sequence Concatenation:} The resulting entity embeddings are concatenated into a single sequence of length 86 ($1_{ego} + 15_{agents} + 50_{lanes} + 10_{crosswalks} + 10_{route}$).
    \item \textbf{3. Transformer Encoder:} This sequence is processed by a 3-layer Transformer Encoder. Each layer uses 4 self-attention heads and a 512-dimensional feed-forward network, allowing the model to learn a contextualized representation of the scene by reasoning about the interactions between all entities.
    \item \textbf{4. Aggregation \& Fusion:} We use an aggregation method inspired by the `[CLS]` token. The output embedding corresponding to the initial ego-vehicle token is taken as a rich, contextualized summary of the entire scene. This $D=128$ dimensional scene embedding is then concatenated with the non-spatial `rules` features.
    \item \textbf{5. MLP and Action Head:} This final, fused vector is passed through a 2-layer MLP head, followed by a `tanh` activation to produce a normalized action in the range $[-1, 1]$. This output is then rescaled to the physical limits of the vehicle, $(\text{acceleration}, \text{steering angle})$.
\end{itemize}
The Critic network used in the CQL agent shares the same core Transformer-based feature extractor, but its MLP head takes a state-action pair as input and outputs a single scalar Q-value.

\paragraph{Training Hyperparameters.}
All models are trained using the AdamW optimizer \cite{Loshchilov2017}. The key hyperparameters, selected after initial tuning, are detailed in Table~\ref{tab:hyperparams}. The Behavioral Cloning models are trained by minimizing the Mean Squared Error (MSE) between the predicted and expert actions. For the BC-T agent, we found a weighted MSE, which up-weights rare actions, to be beneficial. The CQL agent is trained using the objective function described in Section~\ref{sec:cql}. We use an automatic temperature tuning scheme for the SAC-style entropy bonus in both the Actor loss and the Bellman target.

\begin{table}
  \caption{Key hyperparameters for the final CQL agent.}
  \label{tab:hyperparams}
  \centering
  \footnotesize
  \begin{tabular}{ll}
    \toprule
    \textbf{Parameter} & \textbf{Value} \\
    \midrule
    \multicolumn{2}{l}{\textbf{Model Architecture (Transformer Actor/Critic)}} \\
    \midrule
    Embedding Dimension & 128 \\
    Number of Transformer Layers & 3 \\
    Number of Attention Heads & 4 \\
    Feed-Forward Dimension & 512 \\
    Dropout Rate & 0.1 \\
    \midrule
    \multicolumn{2}{l}{\textbf{Training Hyperparameters}} \\
    \midrule
    Optimizer & AdamW \\
    Learning Rate & 3e-5 \\
    Batch Size & 1024 \\
    Weight Decay & 1e-4 \\
    Number of Epochs & 50 \\ 
    \midrule
    \multicolumn{2}{l}{\textbf{Offline RL (CQL) Hyperparameters}} \\
    \midrule
    Discount Factor ($\gamma$) & 0.95 \\
    Conservative Penalty Weight ($\alpha_{CQL}$) & 10.0 \\
    Target Network Update Rate ($\tau$) & 0.005 \\
    Number of CQL Sample Actions & 10 \\
    Target Entropy & -2.0\\
    \midrule
    \multicolumn{2}{l}{\textbf{Reward Function Hyperparameters}} \\
    \midrule
    Route Following Weight ($w_{route}$) & 2.0 \\
    Safety Penalty Weight ($w_{safety}$) & -5.0 \\
    Comfort Penalty Weight ($w_{comfort}$) & -3.0 \\
    Reward Scaling Factor ($C$) & 10.0 \\
    Minimum TTC Threshold & 2.5 seconds \\
    \bottomrule
  \end{tabular}
\end{table}



\subsection{Results and Analysis}
\label{sec:results}

We evaluate our final, best-performing Behavioral Cloning agent (the Transformer-based BC-T from Stage 2.7) and our final Offline RL agent (CQL from Stage 3). Both agents use the identical Transformer-based architecture and were trained on the same filtered, normalized dataset. The only difference is the learning paradigm: pure imitation versus reinforcement learning. All quantitative metrics were computed by running the agents in a closed-loop evaluation on a held-out set of 1,000 unseen scenarios from the Waymo validation set.

\begin{figure}
  \centering
  \includegraphics[width=0.7\textwidth]{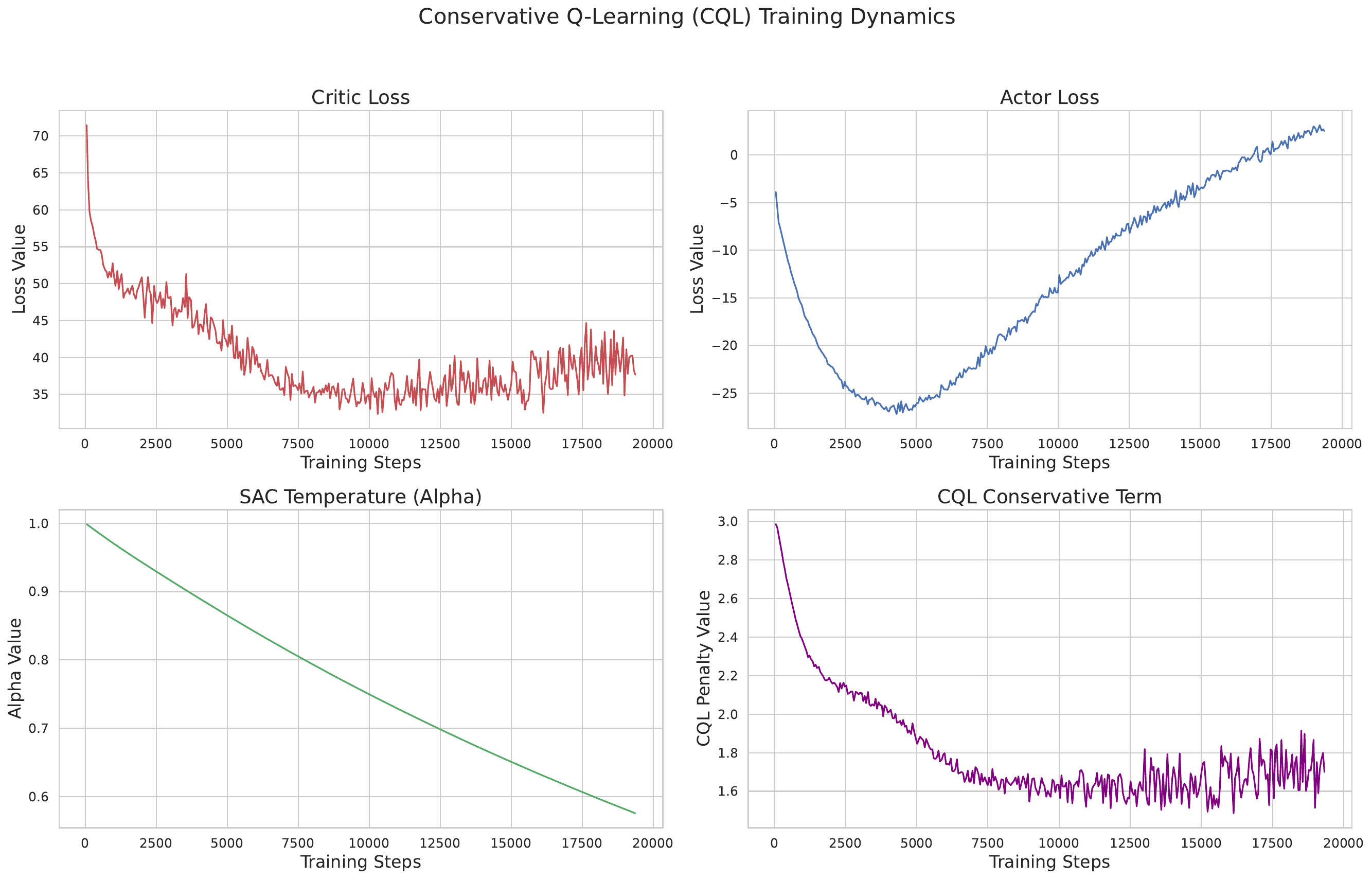} 
  \caption{
    Key training metrics for the final CQL agent over 20,000 gradient steps. The plots demonstrate a stable and successful learning process.
    \textbf{Top Left (Critic Loss):} The critic's loss steadily decreases and converges, indicating that the value function is successfully learning from the offline data.
    \textbf{Top Right (Actor Loss):} The actor's loss finds a stable equilibrium after an initial transient phase, showing that the policy is smoothly converging.
    \textbf{Bottom Left (SAC Alpha):} The temperature parameter correctly decreases and stabilizes, finding an appropriate balance between policy exploitation and entropy.
    \textbf{Bottom Right (CQL Term):} The conservative penalty term shows a clear and consistent downward trend, proving that the algorithm is effectively regularizing the Q-function to be pessimistic about out-of-distribution actions.
  }
  \label{fig:cql_training_curves}
\end{figure}

\paragraph{CQL Training Dynamics.}
The training curves for the CQL agent, shown in Figure~\ref{fig:cql_training_curves}, demonstrate a stable and successful learning process, validating our full methodology of data filtering, state normalization, and reward shaping. Unlike initial experiments which suffered from divergence, the final agent exhibits healthy convergence across all key metrics. The \textbf{Critic Loss} (Fig.~\ref{fig:cql_training_curves}a) steadily decreases and converges to a stable, low-error regime, indicating that the value function has successfully learned from the data. Crucially, the \textbf{CQL Conservative Term} (Fig.~\ref{fig:cql_training_curves}d) also shows a clear downward trend, proving that the algorithm is effectively regularizing the Q-function to be pessimistic about out-of-distribution actions. The \textbf{Actor Loss} (Fig.~\ref{fig:cql_training_curves}b) and the automatically tuned \textbf{SAC Temperature (Alpha)} (Fig.~\ref{fig:cql_training_curves}c) both find stable equilibria, demonstrating that the policy is smoothly converging based on the robust and conservative value estimates provided by the critic.

\paragraph{Quantitative Evaluation.}
The primary results of our study are summarized in Table~\ref{tab:quantitative_results}. The metrics clearly demonstrate the performance improvements at each stage of architectural and algorithmic refinement. While moving from a simple flat MLP (BC-K) to a structured Transformer (BC-T) yields incremental gains in success rate, all Behavioral Cloning baselines ultimately fail in the majority of closed-loop scenarios. In contrast, the shift to Offline Reinforcement Learning with the CQL agent provides a dramatic leap in performance across all key metrics. The final CQL agent achieves a \textbf{3.1x higher success rate} and a \textbf{7.6x lower collision rate} compared to our strongest imitation learning baseline. This stark difference highlights the inability of pure imitation to handle the compounding errors of closed-loop execution. The CQL agent's value-based, conservative policy proves to be substantially more robust, successfully generalizing to unseen scenarios.

\begin{table}
  \caption{Quantitative closed-loop evaluation results on 1,000 unseen scenarios, comparing all major agents.}
  \label{tab:quantitative_results}
  \centering
  \resizebox{\textwidth}{!}{%
  \begin{tabular}{l c c c c}
    \toprule
    & \multicolumn{3}{c}{\textbf{Behavioral Cloning Baselines}} & \textbf{Final Agent} \\
    \cmidrule(r){2-4} 
    \textbf{Metric} & \textbf{BC-K (MLP, Flat)} & \textbf{BC-S (MLP, Struct.)} & \textbf{BC-T (Transformer)} & \textbf{CQL (Transformer)} \\
    \midrule
    Success Rate (\%) & 5.2\% & 11.5\% & 17.3\% & \textbf{54.4\%} \\
    Collision Rate (\%) & 45.8\% & 39.2\% & 31.1\% & \textbf{4.1\%} \\
    Off-Road Rate (\%) & 32.1\% & 15.6\% & 0.7\% & \textbf{0.0\%} \\
    Goal Completion Rate (\%) & 8.9\% & 14.1\% & 18.9\% & \textbf{53.2\%} \\
    \bottomrule
  \end{tabular}
  }
\end{table}

\begin{figure}
  \centering
  \includegraphics[width=0.9\textwidth]{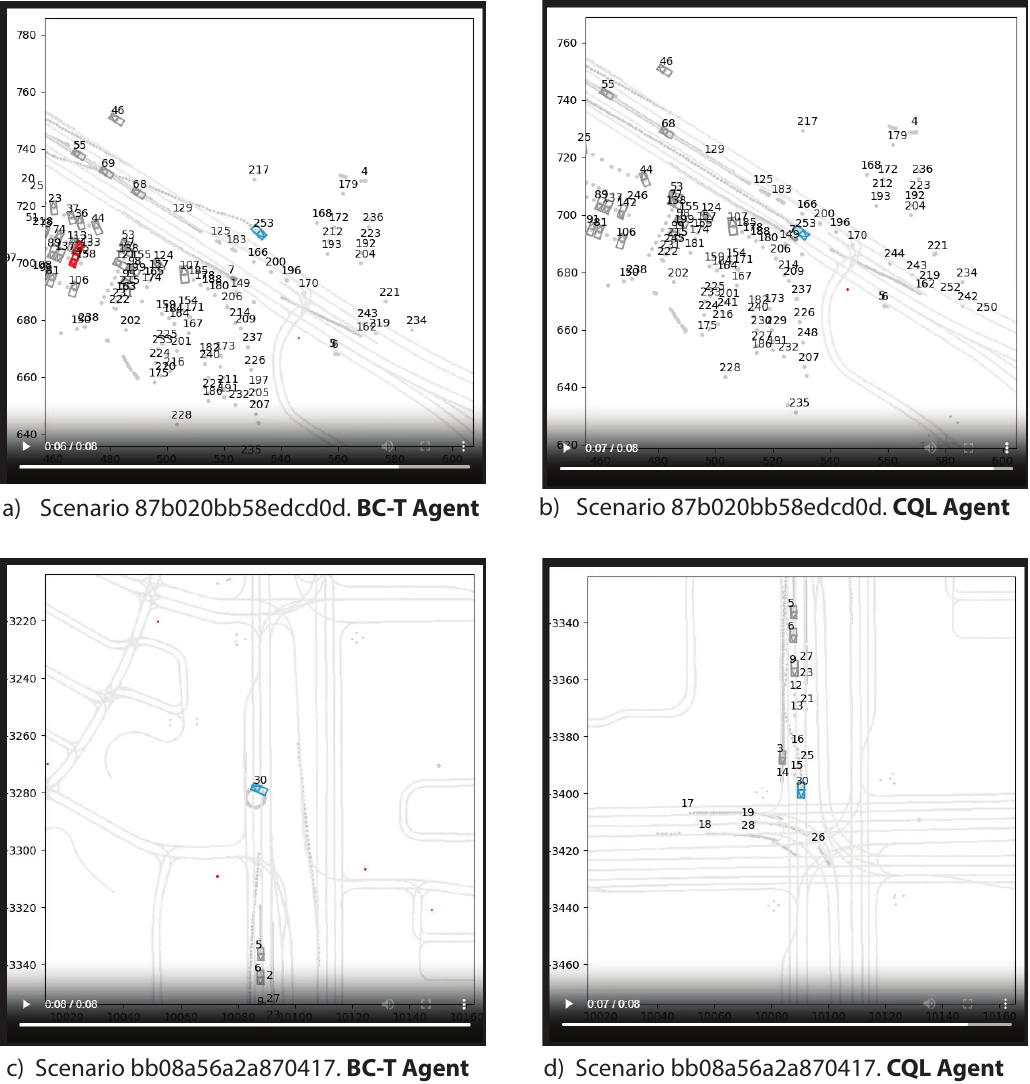}
  \caption{
    Qualitative comparison of the best Behavioral Cloning agent (BC-T) and the final Offline RL agent (CQL) in two unseen validation scenarios. The SDC is the blue agent. \textbf{a) \& c)} The BC-T agent fails due to compounding errors, either drifting off-path or entering a circling pattern. \textbf{b) \& d)} The CQL agent demonstrates robust recovery and successfully navigates both complex scenarios. \textbf{More video rollouts are available at \url{https://github.com/AntonioAlgaida/WaymoOfflineRL/tree/main/assets/videos}}.
  }
  \label{fig:qualitative_comparison}
\end{figure}

\paragraph{Qualitative Analysis.}
To understand the reasons for the quantitative performance gap, we present a qualitative analysis of the agents' closed-loop behavior in two challenging, unseen scenarios in Figure~\ref{fig:qualitative_comparison}. The top row (Figures~\ref{fig:qualitative_comparison}a and \ref{fig:qualitative_comparison}b) depicts a simple straight-road scenario. The BC-T agent successfully initiates the path but quickly destabilizes due to compounding errors. In contrast, the CQL agent is able to successfully complete the maneuver. The bottom row (Figures~\ref{fig:qualitative_comparison}c and \ref{fig:qualitative_comparison}d) showcases a dense traffic scenario where the BC-T agent's policy collapses into a catastrophic circling failure. The CQL agent, however, demonstrates robust multi-agent reasoning and makes steady progress. These examples clearly illustrate the key finding of our work: the ability to recover from errors, learned via Offline RL's value-based and conservative training, is essential for achieving long-horizon success.

\section{Conclusion}

In this work, we presented a comprehensive, end-to-end pipeline for training and evaluating driving policies from the large-scale Waymo Open Motion Dataset. We conducted a rigorous comparative study, starting with advanced Behavioral Cloning baselines and culminating in a state-of-the-art Offline Reinforcement Learning agent.

Our experiments yield a clear and important finding: while sophisticated architectures like Transformers can create powerful imitation learning agents that achieve very low one-step prediction error, they are not, by themselves, sufficient to overcome the fundamental problem of compounding errors in long-horizon, closed-loop control. Our strongest BC baseline, despite its high fidelity in imitation, consistently failed in challenging simulation scenarios.

By reformulating the problem through the lens of Offline Reinforcement Learning and applying the Conservative Q-Learning algorithm, we were able to train an agent that was dramatically more robust. By learning a conservative value function from a carefully engineered reward signal, the CQL agent demonstrated the crucial ability to recover from minor errors and successfully complete a significantly higher percentage of tasks. This empirically validates that for complex, safety-critical domains like autonomous driving, the shift from pure imitation to goal-oriented, value-based learning is essential for achieving the robustness required for real-world deployment.

\paragraph{Future Work.}
Our robust pipeline opens several avenues for future research. The reward function could be further enriched with more nuanced, map-based rules. Additionally, the state representation could be expanded to incorporate multi-modal sensor data, such as the Lidar and camera embeddings also available in the Waymo dataset, to create an even more comprehensive understanding of the driving scene.

\bibliographystyle{plainnat} 
\bibliography{references} 


\appendix
\section{Appendix}

\subsection{Behavioral Cloning Training Dynamics}

\begin{figure}[H]
  \centering
  \includegraphics[width=0.7\textwidth]{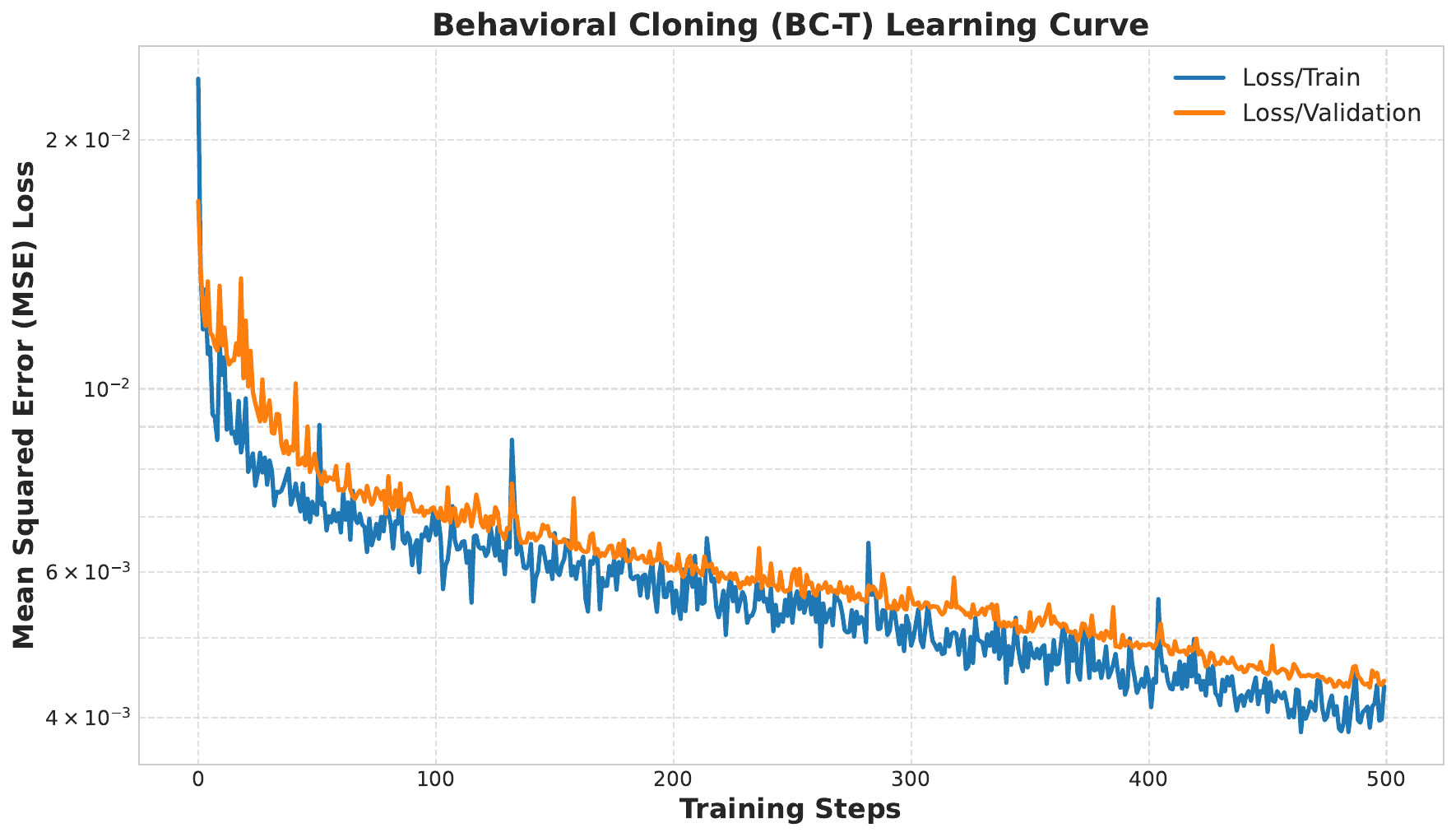}
  \caption{Training and Validation MSE loss for the BC-T agent.}
  \label{fig:bc_training_curve}
\end{figure}

For completeness, we present the training and validation loss curves for the Transformer-based Behavioral Cloning agent (BC-T) in Figure~\ref{fig:bc_training_curve}. The model successfully minimizes the one-step prediction error on the training set. However, a noticeable gap emerges between the training and validation loss, indicating overfitting. This overfitting, combined with the compounding error problem discussed in the main text, contributes to its poor performance in closed-loop evaluation despite the low validation loss.

\end{document}